# Controlling Long-Horizon Behavior in Language Model Agents with Explicit State Dynamics

Sukesh Subaharan[1]

[1]Independent Researcher

Bangalore, India

email: sukeshsubaharan@yahoo.com / sukeshsubaharan05@gmail.com

## ABSTRACT

Large language model (LLM) agents often exhibit abrupt shifts in tone and persona during extended interaction, reflecting the absence of explicit temporal structure governing agent-level state. While prior work emphasizes turn-local sentiment or static emotion classification, the role of explicit affective dynamics in shaping long-horizon agent behavior remains underexplored. This work investigates whether imposing dynamical structure on an external affective state can induce temporal coherence and controlled recovery in multi-turn dialogue.

We introduce an agent-level affective subsystem that maintains a continuous Valence-Arousal-Dominance (VAD) state external to the language model and governed by first- and second-order update rules. Instantaneous affective signals are extracted using a fixed, memoryless estimator and integrated over time via exponential smoothing or momentum-based dynamics. The resulting affective state is injected back into generation without modifying model parameters.

Using a fixed 25-turn dialogue protocol, we compare stateless, first-order, and second-order affective dynamics. Stateless agents fail to exhibit coherent trajectories or recovery, while state persistence enables delayed responses and reliable recovery. Second-order dynamics introduce affective inertia and hysteresis that increase with momentum, revealing a trade-off between stability and responsiveness.

1.      Introduction

Large language models (LLMs) are being widely used as interactive agents in applications such as conversational interfaces, social robots, and autonomous systems that require sustained engagement [1-3]. For these agents to function effectively, they must maintain consistent personality traits and emotional characteristics throughout prolonged interactions spanning multiple conversation turns and sessions. However, research shows that LLM-driven agents frequently demonstrate sudden changes in personality and emotional tone (referred to here as instantaneous affective reversal) which compromise social consistency and diminish user confidence [4-6]. When an agent abruptly switches from showing compassionate understanding to displaying cold indifference within the same conversation, or when it unpredictably fluctuates between different emotional states, such inconsistency disrupts the continuity necessary for successful human-agent interaction [7].

Prior work has attempted to mitigate these issues through various architectural modifications and training strategies. Techniques such as persona-based prompting and role-specific conditioning help steer model outputs toward particular character attributes [8,9], while memory mechanisms including buffer systems and retrieval-augmented memory allow agents to reference conversational history when formulating responses [10-12]. Additionally, instruction-based tuning and model fine-tuning help synchronize agent behavior with target interaction styles. Nevertheless, these methods focus on information storage rather than explicitly governing the evolution of internal states over time [3,13]. They grant agents access to historical context without constraining how emotional transitions unfold temporally. Consequently, agents may still display abrupt emotional discontinuities despite having extensive memory capabilities [14,15].

We contend that the core issue lies not in insufficient memory, but in the lack of inertia. Human emotional states demonstrate temporal stability: feelings do not flip instantaneously but instead change gradually, influenced by both immediate inputs and preceding internal conditions. This stability, known as affective inertia in psychological and dynamical systems literature, indicates that internal states possess momentum and resist sudden changes unless exposed to prolonged or sufficiently intense influences. Building on this principle, we suggest representing affective state as a continuous latent variable controlled by explicit dynamics modeled after physical systems. Instead of treating each response as an isolated output from a stateless process, we incorporate temporal structure into the agent's internal representation, facilitating emotional consistency across interactions.

In practice, we realize this framework using a Valence-Arousal-Dominance (VAD) latent state combined with second-order dynamics that account for both position (present affective state) and velocity (pace of affective transition). Importantly, this method does not require model retraining: the dynamics function as an inference-time overlay around the base LLM, adjusting generation through prompt-based conditioning informed by the external state. The system operates by updating the VAD state via a momentum-driven rule that incorporates affective signals from generated content while honoring inertial boundaries. This architecture ensures compatibility with any pre-trained LLM and permits flexible adjustment of inertia parameters without altering model parameters.

Our experimental results uncover systematic relationships between inertia and affective stability. Baseline agents without hysteresis display unconstrained affective reversals, validating the fundamental problem. Agents configured with moderate inertia show ability to gradually recover from emotional disturbances, smoothly transitioning back to equilibrium states across multiple turns while maintaining sensitivity to legitimate contextual shifts. Conversely, agents with overly strong inertia become locked in particular affective states, unable to adjust when situational changes require emotional recalibration. Notably, although token-level generation retains stochasticity, behavior becomes deterministic at the latent representation level: identical VAD trajectory sequences yield consistent emotional patterns, enabling reproducible and transparent agent dynamics.

These results carry significant implications for developing reliable, socially consistent agents. By incorporating affective inertia, we introduce a coherence mechanism that augments rather than supplants memory-oriented approaches. The resulting systems display smoother emotional transitions, decreased vulnerability to spurious mood changes, and enhanced conformity with human expectations regarding emotional stability. This framework holds particular value for domains such as human-robot interaction, digital mental health interventions, and autonomous agent platforms where sustained social engagement and emotional dependability are essential [16-18]. By anchoring affective dynamics in explicit, adjustable physical principles, we establish a systematic approach toward creating agents that are not merely competent, but also temporally coherent.

## 2. Methods

### 2.1 Experimental Overview

We study whether imposing explicit dynamical structure on an agent's latent affective state can induce temporal coherence and resistance to abrupt persona shifts during multi-turn interaction. Rather than modifying the underlying language model, we introduce an external, low-dimensional latent state governed by physically inspired update rules and evaluate its effect on affective trajectories under adversarial and reconciliatory dialogue.

All experiments follow a fixed, automated multi-turn dialogue protocol (detailed in S1 of Supplementary Material) and compare four conditions: a stateless control, a first-order affective state model, and two second-order momentum-based affective models with differing inertia. Each condition was evaluated across five independent runs with different language model sampling seeds to assess robustness and reproducibility.

Although instantiated here for affective state, the proposed dynamical framework is agnostic to the semantic interpretation of the latent variable and may be applied to regulate other persistent agent-level traits.

## 2.2 Affective State Space

### 2.2.1 Latent Representation

We represent the agent's internal affective state at dialogue turn $t$ as a continuous vector

$$a_t \in \mathbb{R}^3$$

corresponding to the Valence-Arousal-Dominance (VAD) dimensions. This latent state is not part of the language model's hidden activations; instead, it is an explicit, external state variable maintained across dialogue turns. This heuristic mapping is used solely to provide a noisy instantaneous affective signal; no claims are made about psychological fidelity.

The state is initialized at a neutral baseline $a_0 = (0, 0, 0)$. Although the full VAD vector is tracked, our quantitative analysis focuses on the valence component $v_t$, as it most directly reflects affective polarity during adversarial and reconciliatory interaction [19].

### 2.2.2 Instantaneous Affective Signal

At each turn, the dialogue context $x_t$ (including user input and model output) is mapped to an instantaneous affective estimate

$$\hat{a}_t = f(x_t)$$

where $f$ is a fixed affect extraction function and is held fixed across all conditions and runs. This estimate is treated as memoryless and noisy, reflecting the affect implied by the current interaction alone. Temporal coherence is not enforced at this stage.

The instantaneous affective estimate $\hat{a}_t = f(x_t)$ was implemented using a fixed, lexicon-based sentiment analyzer (VADER). The compound sentiment score was used as a proxy for valence, arousal was approximated by sentiment magnitude, and dominance by the difference between positive and negative sentiment scores. This extraction function was held constant across all conditions and trials and introduces no temporal smoothing or memory.

## 2.3 Affective Dynamics

### 2.3.1 Stateless Control Condition

In the stateless baseline, no affective state is maintained across turns. The affective estimate is recomputed independently at each turn:

$$\boldsymbol{a}_t = \hat{\boldsymbol{a}}_t$$

This defines a purely Markovian process with no path dependence or affective memory [20].

### 2.3.2 First-Order Affective Dynamics (No Momentum)

To isolate the effect of state persistence without inertia, we define a first-order update rule:

$$\boldsymbol{a}_{t+1} = \boldsymbol{a}_t + \alpha(\hat{\boldsymbol{a}}_t - \boldsymbol{a}_t)$$

where $\alpha \in (0, 1]$ controls the rate of adaptation; in this study we consider the maximally responsive first-order regime ($\alpha = 1$). This formulation corresponds to exponential smoothing [21] and introduces affective persistence while remaining overdamped and directionally responsive. Although $\alpha = 1$ yields immediate adaptation, the first-order condition differs from the stateless baseline in that the affective state is maintained and fed back into generation across turns.

In experiments, this condition is implemented by maintaining the affective state while setting the momentum coefficient to zero.

### 2.3.3 Second-Order Momentum-Based Affective Dynamics

To model affective inertia, we extend the system to second-order dynamics by introducing an affective velocity

$$\boldsymbol{u}_t = \boldsymbol{a}_t - \boldsymbol{a}_{t-1}$$

Velocity is updated according to

$$\boldsymbol{u}_{t+1} = \mu \boldsymbol{u}_t + (1 - \mu)(\hat{\boldsymbol{a}}_t - \boldsymbol{a}_t)$$

where $\mu \in [0, 1)$ is an inertia (momentum) coefficient. The affective state then evolves as

$$\boldsymbol{a}_{t+1} = \boldsymbol{a}_t + \boldsymbol{u}_{t+1}$$

This formulation is analogous to discretized damped second-order systems in classical mechanics [22] and momentum methods in optimization [23,24]. Larger values of $\mu$ increase resistance to abrupt directional change, producing path-dependent affective trajectories.

*2.3.4 Dynamical Regimes*

This system admits several qualitatively distinct regimes. These include a stateless regime with no persistence or hysteresis to serve as a baseline, a first-order regime ($\mu = 0$) having affective memory without inertia, a second-order regime ($0 < \mu < 1$) where inertial affective dynamics are present with hysteresis and a high-inertia regime ($\mu \to 1$) to simulate excessive persistence. These regimes are explored empirically by varying $\mu$.

To evaluate the impact of affective memory and momentum, the study utilized four experimental conditions: a Stateless Control (no state persistence) to establish a baseline, First-Order Dynamics ($\mu = 0$) providing state persistence without momentum, and two second-order configurations: Moderate Inertia ($\mu = 0.8$) and High Inertia ($\mu = 0.95$) to test balanced and strong momentum, respectively. Each condition was executed across five independent runs ($N = 5$ per condition, 20 runs total) using a standardized 25-turn dialogue protocol. To ensure statistical robustness and variability, each run employed unique independent sampling seeds.

*2.4 Affective State Feedback into Generation*

At each dialogue turn, the current affective state $\boldsymbol{a}_t$ is injected into the model's generation context via a structured natural-language control prompt encoding the valence, arousal, and dominance values. This conditioning influences response tone and discourse style while leaving the underlying language model parameters unchanged. As a result, the system forms a closed-loop interaction in which dialogue content influences affective state evolution, which in turn modulates subsequent generation. To verify that affective state conditioning meaningfully modulates model behavior, we conducted a calibration in which a fixed prompt was generated under varying static VAD settings. Qualitative differences in tone, assertiveness, and conciliatory language were observed across valence levels, confirming that affective state injection has a measurable influence on generation.

*2.5 Dialogue Protocol*

All conditions are evaluated using a fixed, automated multi-turn dialogue protocol designed to induce affective perturbation followed by reconciliation. The protocol consists of a single 25-turn philosophical conversation that remains identical across all trials and conditions (detailed in Appendix A of the Supplementary). No human intervention occurs during evaluation.

As described above, five independent runs were conducted per condition, each with a different language model sampling seed, to verify robustness of the observed affective dynamics across stochastic variations in token-level generation.

*2.6 Recovery Metric*

We define affective recovery as a post-perturbation event rather than a simple threshold crossing.

Let $v_t$ denote the valence component of $\boldsymbol{a}_t$, and let

$$t_{min} = \arg\min_t v_t$$

be the turn at which minimum valence occurs. The recovery turn is defined as

$$t_r = \min\{t > t_{min} \mid v_t > 0\}$$

If no such turn exists within the dialogue horizon, recovery is considered absent.

2.7  *Affective Hysteresis*

To quantify path dependence, we compute affective hysteresis as the area between the valence trajectory during descent and recovery [25,26].

Let $v_\downarrow(t)$ denote the trajectory from the start of the dialogue to $t_{min}$, and $v_\uparrow(t)$ denote the trajectory from $t_{min}$ onward. Hysteresis is approximated via a discrete Riemann sum:

$$H = \sum_t (v_\uparrow(t) - v_\downarrow(t))\Delta t$$

after interpolating the recovery trajectory onto the descent time grid. A zero-hysteresis value indicates a memoryless affective process, while larger values indicate increasing affective inertia and path dependence.

2.8  *Determinism and Reproducibility*

For a fixed dialogue trajectory, affective state evolution is fully deterministic and governed solely by the update equations described above. Token-level text generation remains stochastic due to sampling; repeated trials with independent sampling seeds are therefore used to assess whether macro-scale affective dynamics are robust to stochastic variation in generated text.

2.9  *Language Model and Implementation Details*

All experiments are conducted using a 7B-parameter autoregressive language model quantized to 4-bit precision [27]. Quantization is used to enable efficient experimentation and does not affect the affective dynamics, which operate externally to the model.

For each trial, the full dialogue transcript, per-turn affective state values, and instantaneous affect estimates are logged to disk in structured CSV format. These logs are provided in the Supplementary Materials to enable independent inspection of affective trajectories and associated model outputs.

Since the contribution of this work concerns agent-level state dynamics rather than representational capacity, model size and precision are not expected to qualitatively alter the observed behavior.

Affective state updates, logging, and evaluation are implemented in Python. Each trial logs per-turn affective state values to disk for analysis.

## 3. Results

### 3.1 Valence trajectories exhibit regime-dependent temporal structure

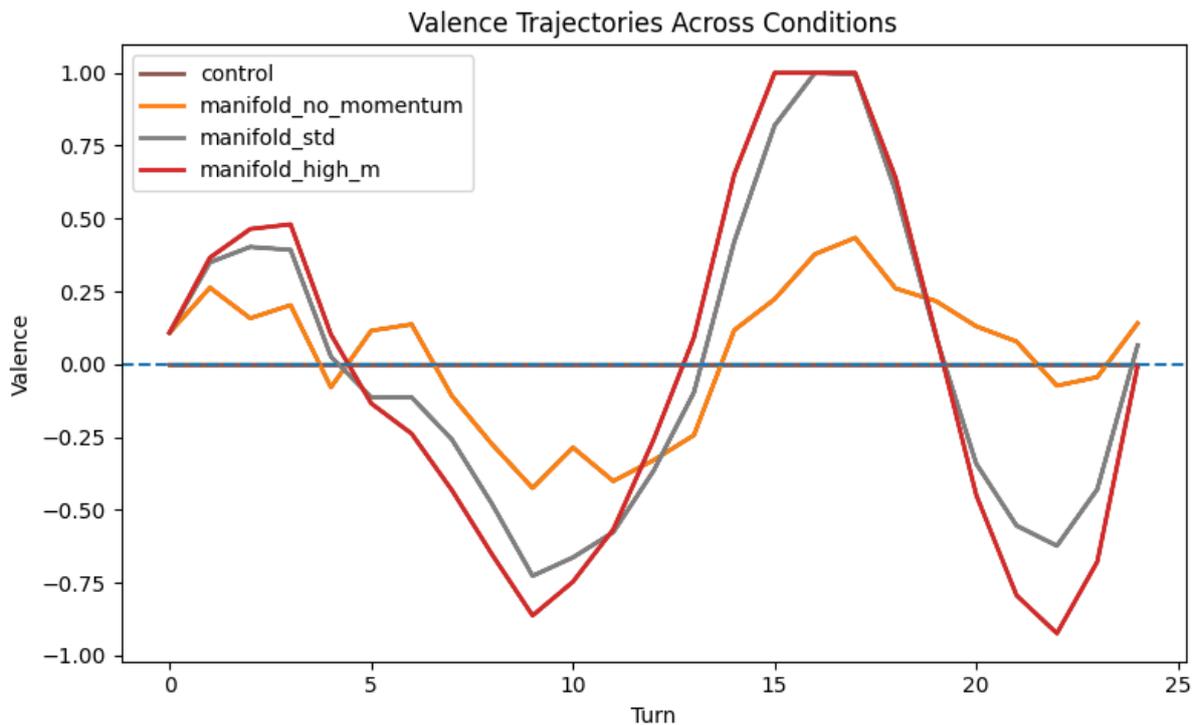

**Figure 1: Mean valence trajectories across affective dynamical regimes.**
Mean valence $v_t$ over a fixed 25-turn dialogue protocol for the stateless control, first-order affective dynamics ($\mu = 0$), and second-order momentum-based dynamics with moderate ($\mu = 0.8$) and high ($\mu = 0.95$) inertia. The adversarial phase begins at turn 6 and the reconciliation phase begins at turn 14. Stateless dynamics exhibit locally reactive but temporally unstructured behavior. Stateful dynamics introduce delayed descent following adversarial onset and recovery aligned with reconciliation. Increasing inertia produces deeper troughs, stronger overshoot, and increased resistance to directional change, with excessive inertia suppressing recovery within the dialogue horizon.

Figure 1 shows the mean valence trajectories across the four affective dynamical regimes under the fixed 25-turn dialogue protocol. The adversarial phase begins at turn 6 and the reconciliation phase begins at turn 14.

In the stateless control condition, valence remains near zero throughout the dialogue and exhibits no coherent descent–recovery structure. Affective responses closely track the instantaneous affective signal, resulting in locally reactive but temporally unstructured behavior.

In contrast, all stateful conditions exhibit a delayed descent in valence following the onset of adversarial interaction. This lag reflects the integration of instantaneous affective estimates into the maintained affective state. The first-order model ($\mu = 0$) produces a smooth but relatively shallow negative excursion, followed by a gradual return toward neutral affect during reconciliation.

Second-order momentum-based models exhibit qualitatively distinct trajectories. The moderate-inertia condition ($\mu = 0.8$) produces a deeper negative trough and a larger positive overshoot during reconciliation, yielding a wider affective loop. The high-inertia condition ($\mu = 0.95$) further amplifies this effect, producing strong resistance to directional change and extended persistence of negative affect.

Across all stateful regimes, the minimum valence occurs several turns after adversarial onset, consistent with the expected temporal lag introduced by affective state dynamics.

*3.2    Recovery behavior aligns with reconciliation onset*

| Condition | Mean Recovery Turn | Recovered (N/5) |
| --- | --- | --- |
| Stateless | ∞ | 0/5 |
| First order ($\mu=0$) | 14 | 5/5 |
| Second order ($\mu=0.8$) | 14 | 5/5 |
| Second order ($\mu=0.95$) | ∞ | 0/5 |

**Table 1: Affective recovery across dynamical regimes.**
Mean recovery turn $t_r$ and number of trials exhibiting recovery within the 25-turn dialogue horizon for each affective regime. Stateless dynamics and high-inertia second-order dynamics fail to recover within the evaluation window, whereas first-order and moderate-inertia dynamics recover consistently at the onset of reconciliation (turn 14).

We next examined affective recovery, defined as the first post-minimum turn at which valence becomes positive. Recovery behavior is summarized in Table 1.

Neither the stateless control nor the high-inertia second-order model recovered within the 25-turn dialogue horizon. In contrast, both the first-order and moderate-inertia ($\mu = 0.8$) models recovered in all trials, with recovery occurring consistently at or around turn 14.

Notably, recovery onset coincides precisely with the reconciliation phase of the dialogue protocol. This alignment suggests that affective state dynamics do not anticipate recovery but instead mediate the system's response to changes in interaction tone. While inertia reshapes trajectory depth and curvature, recovery timing remains governed by external forcing rather than internal momentum alone.

Across independent runs, recovery behavior was highly consistent, indicating that macro-scale affective dynamics are governed primarily by the imposed update rules rather than token-level stochastic variation.

### 3.3 Affective hysteresis increases monotonically with inertia

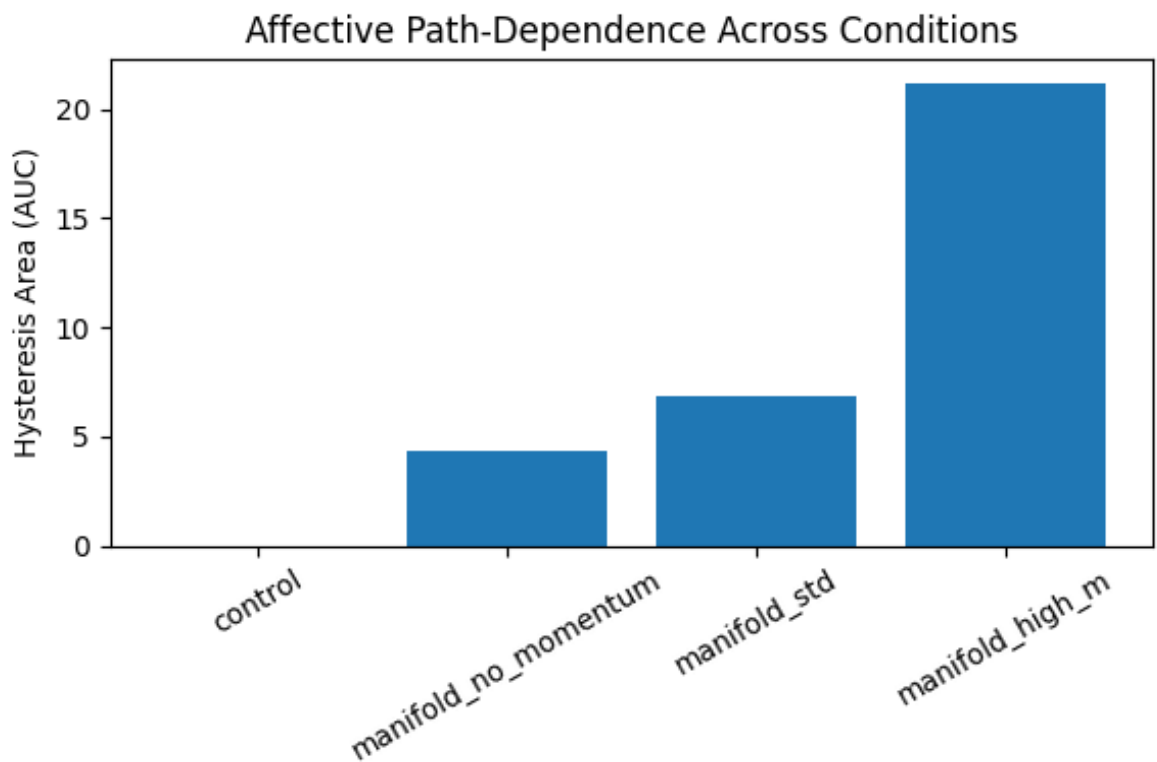

**Figure 2: Affective hysteresis increases with inertia.**
Mean hysteresis area under the curve (AUC) computed between the descent and recovery portions of the valence trajectory for each affective regime. The stateless control exhibits zero hysteresis, reflecting the absence of affective memory. Introducing first-order state persistence yields non-zero hysteresis, while second-order dynamics produce progressively larger hysteresis with increasing inertia. Error bars are omitted due to zero or negligible variation across runs.

To quantify path dependence, we computed affective hysteresis as the area between the descent and recovery portions of the valence trajectory (Figure 2). The stateless control exhibits zero hysteresis by construction, reflecting the absence of affective memory. Introducing first-order state persistence yields a non-zero hysteresis area (AUC = 4.36), indicating the emergence of path-dependent affective behavior. Second-order dynamics further amplify this effect. The moderate-inertia condition ($\mu = 0.8$) exhibits greater hysteresis than the first-order model (AUC = 6.85), while the high-inertia condition ($\mu = 0.95$) produces the largest hysteresis by a substantial margin (AUC = 21.19). Hysteresis increases monotonically with the momentum coefficient $\mu$, demonstrating that affective inertia systematically increases path dependence. These results confirm that second-order dynamics introduce qualitatively distinct affective behavior that cannot be reproduced by first-order smoothing alone.

*3.4    Excessive inertia suppresses recovery within finite horizons*

While affective inertia enhances persistence and hysteresis, excessive inertia introduces a failure mode. The high-inertia condition ($\mu = 0.95$) fails to recover within the dialogue horizon in all trials, despite eventual reconciliation in the dialogue.

This behavior reflects a trade-off between affective stability and responsiveness. Strong inertia confers resistance to abrupt persona shifts but can also impede timely recovery once external conditions change. The moderate-inertia regime balances these competing effects, maintaining persistence while preserving recoverability.

## 4.    Discussion

This work investigates whether imposing explicit dynamical structure on an external affective state can induce temporally coherent and path-dependent behavior in large language model agents during multi-turn interactions. By introducing first and second-order update rules over a low-dimensional Valence-Arousal-Dominance (VAD) state, we show that affective inertia fundamentally reshapes agent-level behavior, producing delayed responses, recovery dynamics and hysteresis that do not arise in stateless or purely reactive systems.

*4.1    Affective dynamics beyond stateless activity*

Prior work in affective computing and sentiment analysis has largely focused on turn-local affect estimation or static emotion classification [28,29]. While such approaches can characterize instantaneous affect, they do not capture temporal structure inherent in dialogue, where emotional responses often lag, accumulate and persist. Our results demonstrate that even a simple first-order state update is sufficient to induce coherent affective trajectories and enable recovery following adversarial perturbation, highlighting the importance of explicit state persistence for modeling agent behavior over time.

The stateless control condition, which recomputes affect independently at each turn, fails to exhibit either recovery or hysteresis, underscoring the limitations of memoryless affective processing. This aligns with previous observations that Markovian dialogue models struggle to maintain consistent persona or affect across extended interactions [20].

*4.2    Second-order inertia and path dependence*

Extending affective state updates to second-order dynamics introduces qualitatively new behavior. Momentum-based updates, inspired by classical damped systems [22] and optimization methods [23,24], produce affective inertia that resists abrupt directional changes and induces path dependence, as quantified by increasing hysteresis. Importantly, this behavior emerges without modifying the underlying language model, suggesting that agent-level dynamics can be shaped through external state evolution alone.

The monotonic increase in hysteresis with the momentum coefficient $\mu$ demonstrates that affective inertia is a controllable property of the system. Moderate inertia produces wider affective loops while preserving recoverability, whereas excessive inertia suppresses recovery within a finite dialogue horizon. This trade-off mirrors stability-responsiveness tensions observed in control-systems and optimization dynamics, where stronger damping improves smoothness at the cost of adaptability.

*4.3    Recovery as a dynamical, not anticipatory, phenomenon*

A key observation is that recovery onset aligns with the reconciliation phase of the dialogue protocol across all recovering regimes. Recovery does not occur prematurely, nor is it accelerated by increased inertia. Instead, inertia reshapes the trajectory depth and curvature while leaving recovery timing governed by external forcing. This suggests that the imposed dynamics mediate how agents respond to changing interaction tone rather than enabling anticipatory or predictive affective shifts. This distinction is important for agent design; where affective inertia can stabilize behavior without decoupling the agent from its environment. Excessive inertia, however, risks over-persistence and delayed adaptation, as observed in the high-$\mu$ regime.

*4.4    Implications for agent stability and design*

Recent work on agentic language models has emphasized memory, planning and tool use, often focusing on representational or architectural modifications. Our findings suggest a complementary approach of regulating agent behavior through explicit, low dimensional-dynamical subsystems that operate independently of the language model's internal representations. Such subsystems offer interpretable control over temporal properties like persistence, lag and hysteresis, which are difficult to enforce reliably through prompt engineering alone. Although instantiated here for affective state, the proposed framework is agnostic to the semantic interpretation of the latent variable. Similar dynamics could be applied to regulate other persistent agent-level traits, such as tolerance, cooperativeness or uncertain sensitivity. This provides a general mechanism for shaping long-horizon behavior in interactive systems.

*4.5    Limitations and future directions*

The study has several limitations. First, affect extraction relies on a fixed lexicon-based sentiment estimator, which provides only a coarse proxy for affective state. While the comparative results are robust to extractor noise, richer affect modeling could further refine trajectory shape. Second, experiments were conducted under a single, fixed dialogue protocol. This may limit the claims about generalization across interaction styles or domains. Finally, the evaluation horizon of the study was finite. Longer dialogue may reveal additional dynamical regimes, including eventual recovery in high-inertia regimes.

Future work could explore adaptive or state-dependent inertia, multidimensional analysis beyond valence and integration with learned or task-conditioned affect extractors. Extending these dynamics to multi-agent settings or coupling them with planning and memory modules may further clarify the role of explicit state-based evolution in stabilizing agent behavior.